\documentclass[10pt,twocolumn,letterpaper]{article}
\usepackage{iccv}
\makeatletter
\@namedef{ver@everyshi.sty}{}
\makeatother
\usepackage{tikz}
\usepackage{times}
\usepackage{epsfig}
\usepackage{graphicx}
\usepackage{amsmath}
\usepackage{amssymb}
\usepackage{pgfplots}
\usepackage{latexsym}

\usepackage{booktabs}
\usepackage{microtype}   
\usepackage{nicefrac}  
\usepackage{algpseudocode}
\usepackage{algorithm}
\usepackage{pdfrender}
\usepackage{color}
\usepackage{multirow,graphicx,paralist,bigdelim}
\usepackage{colortbl}
\usepackage{tabularx}
\usepackage{enumitem}
\usepackage{listings}
\usepackage{soul}

\newcommand{\blue}[1]{\textcolor{black}{#1}}

\iccvfinalcopy 

\def\iccvPaperID{8328} 

\ificcvfinal\fi

\usepackage[pagebackref=true,breaklinks=true,letterpaper=true,colorlinks,bookmarks=false]{hyperref}

\begin{document}

\title{Global Pooling, More than Meets the Eye: \\ Position Information is Encoded Channel-Wise in CNNs}


\author{Md Amirul Islam\thanks{Equal Contribution} \:$^{1,6}$ \hspace{0.2cm} Matthew Kowal\footnotemark[1] \:$^{2,6}$ \hspace{0.2cm} Sen Jia$^{4}$\hspace{0.2cm} \text{Konstantinos G. Derpanis}$^{2,5,6}$\hspace{0.2cm} \text{Neil D. B. Bruce}$^{3, 6}$ \\
$^1$Ryerson University, Canada \hspace{0.3cm} 
$^2$York University, Canada \hspace{0.3cm} 
$^3$University of Guelph, Canada \\

$^4$Toronto AI Lab, LG Electronics \hspace{0.3cm}
$^5$Samsung AI Centre Toronto, Canada \hspace{0.3cm}
$^6$Vector Institute for AI, Canada\\ 
\tt\small \text{amirul@cs.ryerson.ca}, \text{\{m2kowal,kosta\}@eecs.yorku.ca}, \text{sen.jia@lge.com}, \text{brucen@uoguelph.ca}
}

\maketitle
\ificcvfinal\thispagestyle{empty}\fi

\begin{abstract}
In this paper, we challenge the common assumption that collapsing the spatial dimensions of a 3D (spatial-channel) tensor in a convolutional neural network (CNN) into a vector via global pooling removes all spatial information. Specifically, we demonstrate that positional information is encoded based on the ordering of the channel dimensions, while semantic information is largely not. Following this demonstration, we show the real world impact of these findings by applying them to two applications. First, we propose a simple yet effective data augmentation strategy and loss function which improves the translation invariance of a CNN's output. Second, we propose a method to efficiently determine which channels in the latent representation are responsible for (i) encoding overall position information or (ii) region-specific positions. We first show that semantic segmentation has a significant reliance on the overall position channels to make predictions. We then show for the first time that it is possible to perform a `region-specific' attack, and degrade a network's performance in a particular part of the input. We believe our findings and demonstrated applications will benefit research areas concerned with understanding the characteristics of CNNs. Code is available at: \href{https://github.com/islamamirul/PermuteNet}{https://github.com/islamamirul/PermuteNet}.
\end{abstract}


\section{Introduction}
One of the fundamental ideas behind different neural network architectures~\cite{he2016deep,Simonyan14,szegedy2015going,chen2018deeplab,he2017mask} is the idea of \textit{invariance}. Given an input signal, an \textit{X}-invariant operation is one that produces the same output regardless of any change (of some type \textit{X}) to the input. This property is desirable in a multitude of applications in computer vision, the most obvious being object recognition~\cite{krizhevsky2012imagenet,he2016deep,Simonyan14}; the goal being to assign the corresponding image-level label (e.g., \textit{dog}) regardless of where the object is located in the image. This is referred to as \textit{translation} invariance. Another property of operations that is intimately related to translation invariance is that of translation \textit{equivariance}: shifting the input and then passing it through the operation is equivalent to passing the input through the operation, and then shifting the signal.

To achieve invariant neural networks, a common strategy is to use \textit{equivariant} operations on a per-layer basis~\cite{cohen2016group}, which then culminates in an \textit{invariant} output. One of the best examples of this can be found in convolutional neural networks (CNNs) for the task of image classification. Following a hierarchy of translation equivariant convolutional layers, CNNs use a global pooling layer to transform the 3D (spatial-channel) tensor into a 1D vector, which is then fed into a fully connected layer to produce the classification logits. One would therefore (intuitively) assume that after the spatial dimensions are collapsed due to the global pooling operation~\cite{lin2013network, szegedy2015going}, spatial information should be removed while translation invariance is produced. However, previous work has shown that absolute position information exists both in the latent representations~\cite{islam2020much} as well as the output of the network~\cite{kayhan2020translation, islam2021position}.

None of these previous works have answered the critical question: How can a CNN contain positional information in the representations if there exists a \textit{global pooling layer} in the forward pass? In this paper, we provide an answer to this question, and demonstrate through rigorous quantitative experiments that CNNs do this by encoding the position information along the \textit{channel} dimension even though the spatial dimensions are collapsed. Moreover, we show that the \textit{position} information is encoded based on the \textit{ordering of the channel dimensions}, while the \textit{semantic} information is largely invariant to this ordering. We argue that these findings are important to better understand the properties of CNNs and to guide their future design.

To demonstrate the real-world impact of these findings, we leverage the fact that position information is encoded channel-wise in a number of domains and applications. First, we tackle the lack of translation invariance of CNNs. We propose a simple yet effective loss function which minimizes the distance between the encodings of translated images to achieve higher translation invariance. Second, we propose an efficient way to identify which channels in the latent representation are responsible for encoding both (i) position information in the entire image and (ii) `region-specific' position information (e.g., channels which activate for the left part of the image). We show quantitative and qualitative evidence that networks have a significant reliance on these channels when making predictions compared with randomly sampled channels. Finally, we show it is possible to target region-specific neurons, and impair the performance in a particular part of the image. To summarize, our main contributions are as follows:

\begin{itemize}
    \item We reveal how global pooling layers admit spatial information via an \textit{ordered} coding along the \textit{channel} dimension. We then demonstrate the real-world applicability of this finding by applying it to the problem domains that follow in this list.
    \item We propose a simple data-augmentation strategy to improve the translation invariance of CNNs by minimizing the distance between the encodings of translated images.
    \item We propose a simple and intuitive technique to identify the position-specific neurons in a network's latent representation. We show that multiple complex networks contain a significant reliance on these position-encoding neurons to make correct predictions.
    \item We show it is possible to attack a network's predictions in a region-specific manner, and demonstrate the efficacy of this approach on a standard self-driving semantic segmentation dataset.
\end{itemize}

\begin{figure*} [t]
\centering
  \begin{center}
  \resizebox{0.99\textwidth}{!}{ 
      \includegraphics[width=0.95\textwidth]{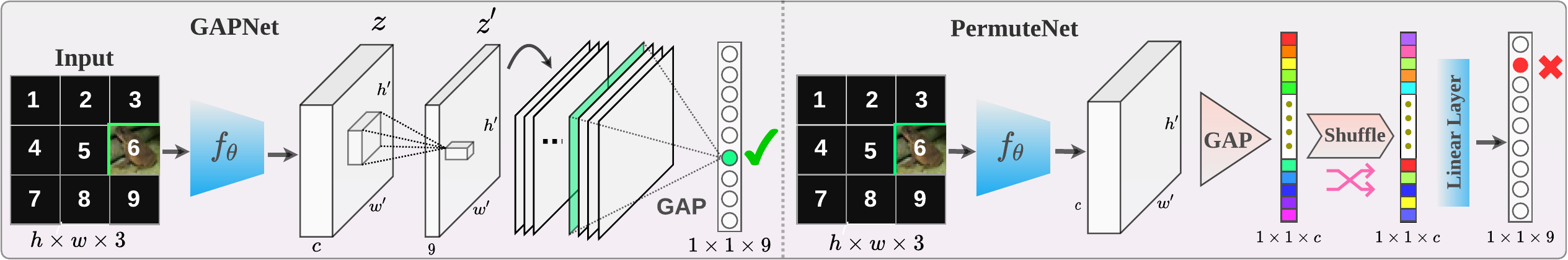}}
  \end{center}
  \vspace{-0.3cm}
  \caption{An illustration of our GAPNet (left) and PermuteNet (right) architectures used to determine the existence of channel-wise positional encodings in CNNs. \textbf{Left:} We feed a grid based input image to the encoder, $f_\theta$, of standard CNNs (e.g., ResNet-18~\cite{he2016deep}) to obtain a latent representation, $z$. $z$ is then transformed to a representation, $z^\prime$, through the last convolutional layer which has the output channel dimension set to the number of locations in the input grid (e.g., 9 in the above example). This enforces the global average pooling (GAP) layer to output the number of locations. The network is then trained to predict the location of the image patch. \textbf{Right:} PermuteNet follows the same structure of a standard CNN except we shuffle the dimensions of the latent representation to verify whether obfuscating the channel ordering hurts the positional encoding capacity.}
  \label{fig:gap_shuffle}
\end{figure*}
\section{Related Work}
Islam et al.~\cite{islam2020much} first demonstrated that \textit{absolute} positional information is captured in a CNN's latent representations. More specifically, they took a frozen pre-trained CNN, and then trained a read-out module to predict a gradient-like position map. They showed absolute position information (e.g., pixel coordinates) could be extracted from many stages of the CNN and that \textit{zero padding} was a significant cause of this information to be encoded. The ability of zero padding to inject position information has been an area of focus in multiple follow-up works~\cite{kayhan2020translation,alsallakh2020mind, islam2021position,liu2018partial,liu2018intriguing}. Kayhan et al.~\cite{kayhan2020translation} demonstrated a number of properties concerning zero padding and positional information, including sample efficiency and generalization to out-of-distribution locations. 
Mind the pad~\cite{alsallakh2020mind} explored zero  padding with respect to object recognition, and demonstrated that it causes severe artifacts in intermediate representation.
Finally, Islam et al.~\cite{islam2021position} performed a large-scale case study on border heuristics, including padding and canvas colors, and showed that positional information can hurt (e.g., texture recognition~\cite{cimpoi2014describing}) or help (e.g., semantic segmentation and instance segmentation~\cite{wang2020solo}) performance depending on the task. \blue{\textit{Relative} position information refers to the position of features in relation to one another within an image. One example of how relative position is used in the design of computer vision algorithms is in capsule networks~\cite{sabour2017dynamic}, which encode every object with a pose (i.e., translation and rotation). This encoding results in later layer features which activate only for specific relative positions from earlier layers. We view relative position information as a different line of research compared with absolute position and thus focus our analysis on solely the latter in this paper.
}

Gatys et al.~\cite{gatys2015texture} first mentioned absolute position in the latent representations of CNNs in a footnote which comments on how zero padding effects generative modelling. 
A more recent study explored zero padding in generative models in a much deeper and systematic way and showed that CNNs use it to produce more robust spatial structure~\cite{Xu_2020_Pos}. 

A handful of recent studies showed that CNNs are not fully translation invariant. For example, BlurPool~\cite{zhang2019making} demonstrated that small pixel-level shifts in the input produce large fluctuations in the output classification probabilities. To make CNNs more invariant to translations, they proposed adding a Gaussian blurring layer after each max-pooling layer within a CNN, which significantly improved the translation invariance of the network. Zou et al.~\cite{zou2020delving} extended this work by using content-aware learned filter weights, which predict separate filter weights for each spatial and channel location in the input. While the proposed solution is effective, BlurPool~\cite{zhang2019making} does not explain the underlying mechanisms of \textit{why} CNNs are not translation invariant in the first place. In our paper, we extend beyond the previous works which showed the existence of position information in CNNs, by explaining for the first time the precise mechanism which allows CNNs to encode position information despite the global pooling layer. 
\section{Channel-wise Position Encoding in CNNs}\label{sec:channel_position}
Recent works~\cite{islam2020much,kayhan2020translation,alsallakh2020mind,islam2021position} showed that CNNs exploit absolute position information. However, no efforts have identified the mechanism in which position information is encoded after global average pooling (GAP) layers. Given the apparent importance of position information, one might raise the question of whether spatial information is being retained by some means. In this section, we aim to answer this interesting question through a series of experiments. We show that absolute position information, despite the spatial dimensions being collapsed, can encode channel-wise after a global pooling layer, in the $1 \times 1 \times C$ latent representation. 

\begin{center}
    \begin{table}
    \def\arraystretch{1.15}
   \setlength\tabcolsep{4.3pt}
        \centering
        \resizebox{0.49\textwidth}{!}{
    \begin{tabular}{cc ccc c ccc} 
             \specialrule{1.2pt}{1pt}{1pt}
              \multicolumn{1}{c}{\multirow{2}{*}{Padding}}  &\multicolumn{1}{c}{\multirow{2}{*}{Network}}  & \multicolumn{3}{c}{Loc. Cls. Acc (\%)} && \multicolumn{3}{c}{Image Cls. Acc (\%)}\\
             \cline{3-5} \cline{7-9}
        
             & & 3$\times$3 & 5$\times$5 & 7$\times$7 && 3$\times$3 & 5$\times$5 & 7$\times$7  \\
           \specialrule{1.2pt}{1pt}{1pt}
            
            \multirow{2}{*}{\textit{Zero}} &GAPNet & 100 & 100 & 100 & & 82.6 & 82.4& 82.1 \\
            
            &PermuteNet & 78.8 & 37.8 & 21.4 &  &  73.6 & 72.2& 69.9\\
            
            \midrule
            
             \multirow{2}{*}{\textit{Reflect}} &GAPNet & 100 & 100 & 100 & & 83.8 & 83.4& 82.9 \\
            
            &PermuteNet & 78.3 & 36.3 & 23.2 &  &  71.1 & 71.4& 65.7\\
            
            \midrule
            
             \multirow{2}{*}{\textit{Replicate}} &GAPNet & 100 & 100 & 100 & & 83.1 & 82.9& 82.8 \\
            
            &PermuteNet & 78.3 & 40.1 & 23.6 &  &  72.0 & 71.5& 71.4\\

            \specialrule{1.2pt}{1pt}{1pt}
        \end{tabular} }
        \caption{Performance comparison between \textit{GAPNet} and \textit{PermuteNet} in terms of absolute location classification and object recognition under various padding types. Regardless of the padding, it is clear that \textit{GAPNet} achieves 100\% positional encoding accuracy while \textit{PermuteNet} fails to correctly classify locations. This demonstrates that positions are encoded channel-wise in the latent representation. Object recognition accuracy further confirms that position information depends on the ordering of the channels while semantic information does not heavily rely on ordering. }
        \label{tab:gap_shuffle}
    \end{table}
\end{center}

\vspace{-1.4cm}
\subsection{Learning Positions with a GAP Layer}
Existing \textit{K-class} object recognition networks~\cite{lin2013network,szegedy2015going,he2016deep} largely follow a similar structure including a feature encoder network followed by a classifier. Given an input image, $I\in \mathbb{R}^{h\times w \times 3}$, the encoder network, $f_\theta$, maps the $h \times w \times 3$ input to a latent representation, $z\in h^\prime \times w^\prime \times c$ where $h' < h$, $w' < w$, and $c \gg 3$. The latent representation, $z$, is then passed to a global average pooling~\cite{lin2013network} operator which collapses the spatial dimensions of $z$ and outputs $\hat{z} \in 1 \times 1 \times c$ representation. This 1-D representation is finally passed to a linear classification layer which produces the $K$ categorical logits, a set of un-normalized scores representing the likelihood of the object class, $k \in K$ being the label of the current image. As the 1-D representation, $\hat{z}$, has no spatial dimension, the common assumption is that position information is certainly lost. 

To demonstrate that the GAP operation can retain absolute positional information of an object, we design two network architectures which we term GAPNet and PermuteNet (see Fig.~\ref{fig:gap_shuffle}). GAPNet follows a similar structure to a standard CNN (e.g., ResNet-18~\cite{he2016deep} and NIN~\cite{lin2013network}) for object recognition, except we remove the final fully connected layer, such that the last layer of the network is the GAP layer. To ensure the GAP layer outputs the correct number of classes, $K$, we set the number of output channels of the last convolutional layer (i.e., the layer before the GAP layer) to $K$. Formally, the last convolutional layer takes the latent representation, $z \in h^\prime \times w^\prime \times c$ as input and outputs a representation, $z^\prime \in h^\prime \times w^\prime \times K$. The intuition of removing the fully connected layer is to enable the GAP layer to predict the class logits. The output of the GAP layer will thus be the same size as the categorical logits and can be used as the last layer of the network (see Fig.~\ref{fig:gap_shuffle} left). 


PermuteNet also follows the structure of a standard object classification network, except for a single \textit{shuffle} operation which occurs between the GAP layer and the penultimate linear layer. This operation randomly shuffles the channel indices of the representation of the GAP layer, which is then passed to the linear layer for classification (see Fig.~\ref{fig:gap_shuffle} right). Note that we construct GAPNet to clearly demonstrate that the output of a GAP layer can directly map to specific absolute positions in the input. We design PermuteNet to show that randomly shuffling the channel ordering hinders the network's ability to encode position information. 

\subsection{Evaluation of Channel-wise Position Encoding}
To validate the existence of a channel-wise position encoding, we design a simple location-dependant task using GAPNet and PermuteNet, such that the output logits can be directly mapped to specific locations in the input image. Inspired by previous works~\cite{kayhan2020translation,islam2021position}, we first conduct a location classification experiment, where each input is a CIFAR-10~\cite{krizhevsky2014cifar} image placed on a $n \times n$ grid, where each of the pixels not containing the image is set to \textit{zero} (see \textit{input} in Fig.~\ref{fig:gap_shuffle}). The target for each input is the location where the image patch is placed (e.g., the target for the given input in Fig.~\ref{fig:gap_shuffle} is 6). We use a ResNet-18~\cite{he2016deep} architecture to report experimental results for GAPNet and PermuteNet under three different padding types. For the location classification task, we train GAPNet and PermuteNet for 20 \textit{epochs} with a learning rate of 0.001, and use the ADAM optimizer~\cite{kingma2014adam}. The number of output logits is set to the number of input locations (e.g., 81 for a $9 \times 9$ grid). We also train the networks for object recognition using grid based data settings by changing the number of output logits to 10 for both GAPNet and PermuteNet. For the location-dependent object classification task, we train GAPNet and PermuteNet for 100 \textit{epochs} with a learning rate of 0.01. \\

\begin{table} [t]
\def\arraystretch{1.15}
\setlength\tabcolsep{3.4pt}
\centering

\resizebox{0.49\textwidth}{!}{

	\begin{tabular}{l cccc cccc ccccc}

	\specialrule{1.2pt}{1pt}{1pt}
	
	\multirow{2}{*}{\rotatebox{0}{Methods}} & &\multicolumn{2}{c}{\text{CIFAR-100}~\cite{krizhevsky2014cifar}} & &\multicolumn{2}{c}{\text{CIFAR-10}~\cite{krizhevsky2014cifar}} && \multicolumn{3}{c}{\text{ImageNet}~\cite{ImageNet}}\\ 
	 
	 \cline{3-4} \cline{6-7} \cline{9-11} 
	 
	 & & \multicolumn{1}{c}{Top-1} & \multicolumn{1}{c}{Cons.8} && \multicolumn{1}{c}{Top-1} & \multicolumn{1}{c}{Cons.8} &&
	 \multicolumn{1}{c}{Top-1 } & \multicolumn{1}{c}{Cons.8} & \multicolumn{1}{c}{Cons.16} \\
	 
	 \specialrule{1.2pt}{1pt}{1pt}
	  
	 ResNet-18~\cite{he2016deep} && 72.6  & 70.1 && 93.1 & 90.8 && 69.7 & 89.5 &  87.4\\
	 +AugShift (ours) && 72.6 & 85.6 && 92.1 & 94.8 && 70.1 & 90.2 & 88.2\\
	 \midrule
	 
	 Blurpool~\cite{zhang2019making} &&  72.4 & 78.2 && 92.5 & 92.5 && 71.4 & 90.5 & 88.8  \\
	 
	 
	 

	   
	\specialrule{1.2pt}{1pt}{1pt}

	\end{tabular}
	
	}
\caption{Comparison of overall classification accuracy and shifting consistency (Cons.) for different networks. Interestingly, our simple \textit{AugShift} technique outperforms the baselines on a smaller dataset (e.g., CIFAR-10 and CIFAR-100) while achieving competitive performance on the large-scale ImageNet dataset. Cons.8 and Cons.16 refer to randomly shifting the input horizontally and vertically by up to 8 and 16 pixels, respectively.}

    \label{tab:invariance}
    \vspace{-0.5cm}
\end{table}

\noindent \textbf{Results.} We present the location classification and object recognition results of GAPNet and PermuteNet in Table~\ref{tab:gap_shuffle}. For the location classification task, GAPNet achieves a counterintuitive \textit{100\% accuracy for all tested grid sizes}. It is clear that the GAP layer can admit robust positional information which directly represents absolute locations on the input image. In contrast, while PermuteNet can learn to identify a marginal number of positions for small gridsizes (e.g., $3 \times 3$), the shuffling of the channel dimension significantly degrades the ability of the network to perform location classification as the grid size increases. This provides direct evidence that the \textit{order of the channel dimensions} is the main representational capacity which allows for the GAP layer to admit absolute position information.
We further evaluate GAPNet and PermuteNet for an image recognition task using similar data settings and report the results in Table~\ref{tab:gap_shuffle} (right). Interestingly, unlike the location classification task, PermuteNet can achieve classification performance close to GAPNet. This reveals an interesting dichotomy between the type of coding CNNs use for \textit{positional} and \textit{semantic} representations: position information depends mainly on the \textit{ordering} of the channels, while semantic information does not.

We show in these experiments that GAP layers can admit position information through the ordering of the channel dimensions. Another interesting question we explore is how much position information can be decoded from \textit{pre-trained} models which are not trained explicitly for location classification. Towards answering this question, we provide more experiments in the supplementary and evaluate the amount of positional information contained in the channel-wise ordering of networks trained for various tasks, such as image classification~\cite{he2016deep} and semantic segmentation~\cite{long15_cvpr}. In further support of our hypothesis, these results also demonstrate unequivocally that GAPNet can recover spatial position information while PermuteNet can not. 


\section{Applicability of Channel-Wise Positional Encoding}

We now demonstrate multiple ways in which our finding that positional information is encoded channel-wise can be leveraged. To ensure that the channel-wise encoding is the source of improvement for each of these applications, we use the representation after passing it through a GAP layer in each case. First, we propose a simple loss function to improve translation invariance in CNNs. Next, we explore the robustness of object recognition networks. We demonstrate different ways to \textit{target} these models using the implicit positional encodings (i.e., as an adversary) both for overall performance and for a region-specific attack.

\begin{figure}
\centering
  \begin{center}
  \resizebox{0.49\textwidth}{!}{ 
      \includegraphics[width=0.98\textwidth]{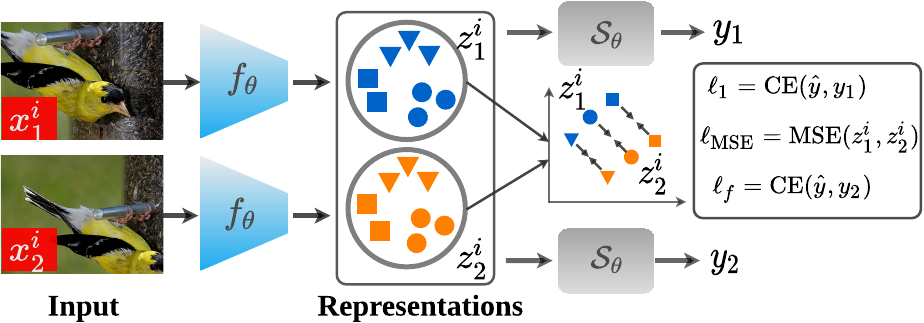}
      }
  \end{center}
  \vspace{-0.35cm}
  \caption{\textcolor{black}{Illustration of the overall training pipeline of our proposed translation invariant model. We use two different crops of the input image, $x^i$ to generate $x^i_1$ and $x^i_2$, which are then both passed to a convolutional encoder network, $f_\theta$, to obtain latent representations, $z^i_1$ and $z^i_2$, respectively. We minimize the distance between the two latent representations, $z^i_1$ and $z^i_2$, through the mean squared error loss which enforces the model to reduce positional bias. The latent representations are also passed to a classification head, $\mathcal{S}_\theta$, which produces the class logits, $y_1$ and $y_2$, respectively.}}
  \label{fig:augshift}
  \vspace{-0.2cm}
\end{figure}
\subsection{Translation Invariance for Object Recognition}

We first tackle the issue of CNNs not being translation invariant. More specifically, recent work~\cite{zhang2019making} showed that small shifts in the input can cause large shifts in the output class likelihoods and proposed BlurPool~\cite{zhang2019making} which applies a Gaussian kernel with every pooling layer of a network. 




	
	 
	 
	 
	  
	 
	 
	 
	 



	


\noindent \textbf{Learning Translation Invariant Representations.} A truly shift invariant network should produce the same output logits, regardless of its shift (i.e., a \textit{cat} is a \textit{cat}, regardless of its \textit{position} in the image). Given that positional information is encoded in the latent representation \textit{preceding} the output logits (i.e., the representation after the GAP layer but before the linear layer), we propose to minimize the difference in this representation between different shifts of the same image. The overall training pipeline is illustrated in Fig.~\ref{fig:augshift}. Given two different crops of the same image, denoted as $x^i_1$ and $x^i_2$, where $i$ is the image index and the subscript denotes the different crops, we feed these images through the same encoder network, $f_\theta$, and obtain the latent representations, $z^i_1$ and $z^i_2$, after the \textit{GAP} layer. We enforce them to be shift invariant by minimizing the distance between the latent representations using the mean squared error (MSE) as a loss term. The latent representations are then passed through the prediction layer, $\mathcal{S}_\theta$, which generates the class logits, $y_1$ and $y_2$, respectively. We apply two cross entropy losses, $\ell_{1}$ and $\ell_2$ between $\{y_1, \hat{y}\}$ and $\{y_2, \hat{y}\}$, respectively where $\hat{y}$ is the object class. Our overall loss function, $L_{\text{AugShift}}$, is then defined as:

\begin{equation}
L_{\text{AugShift}} = \ell_{1} + \ell_{2} + \lambda \cdot \ell_{\text{MSE}} (z_1^i, z_2^i) \ ,
\end{equation}

\noindent where $\lambda$ refers to the loss weight which controls the contribution of MSE loss. \\

\noindent \textbf{Shift Invariance and Accuracy.} To validate the effectiveness of our proposed training strategy, we show overall performance and shift consistency results on the CIFAR-10~\cite{krizhevsky2014cifar}, CIFAR-100~\cite{krizhevsky2014cifar}, and ImageNet~\cite{ImageNet} datasets. We adopt the same consistency (Cons.) metric as proposed previously~\cite{zhang2019making} to measure the translation invariance. More specifically, we measure how often a network predicts the same category after the input image is vertically and horizontally shifted a random number of pixels: up to 8 pixels (Cons.8) for CIFAR, and 8 or 16 (Cons.16) pixels for ImageNet (results for additional shift magnitudes are reported in the supplementary. 

The classification and consistency results are presented in Table~\ref{tab:invariance}. Our approach achieves competitive Top-1 accuracy on CIFAR-10 compared to the baseline but significantly outperforms in terms of shifting consistency (94.8\% vs.\ 90.8\%). Our approach matches the overall classification accuracy on CIFAR-100 while significantly improving the shifting consistency (85.6\% vs.\ 70.1\%). Note that our approach admits greater shift consistency than the BlurPool on the CIFAR datasets. Our method improves the overall performance and the shift consistency on ImageNet compared to the ResNet baseline and remains competitive in terms of performance and consistency with~\cite{zhang2019making}. We believe the performance gap is less (which is true for AugShift and BlurPool) for larger datasets with high resolution (e.g., ImageNet) because the network can learn some degree of translation invariance from the data without any invariance-specific regularization, as proposed. Note that~\cite{zhang2019making} must modify the majority of layers in the CNN architecture by adding another filter at each pooling layer, which in turn adds additional computations at inference time. In contrast, ours is architecture agnostic and adds no computational overhead during inference. \blue{Also our motivation for achieving translation invariance differs from Blurpool (i.e., enforcing the similarity of channel-wise position information vs.\ anti-aliasing) and thus the two techniques may be complementary.} 

\subsection{Attacking the Position-Encoding Channels}\label{sec:attack_channel}
We now aim to demonstrate that complex networks trained for position-based tasks, such as semantic segmentation~\cite{chen2018deeplab,islam2017gated}, have a significant reliance on the position information encoded channel-wise in their latent representations. To perform this type of attack, we first propose a simple and intuitive technique to estimate the position encoding neurons in a CNN's latent representations. A simple modification of the technique allows us to identify the channels which largely encode: (i) overall position or (ii) region-specific position. Then, we evaluate the network's reliance on them for making predictions by simply turning them off during inference. We show that the removal of these neurons causes significantly more harm to the performance than randomly sampled neurons. This suggests a significant encoding of positional information is contained within them. We additionally show the feasibility of performing region-specific attacks on networks trained for autonomous driving tasks. These results demonstrates that channel-wise positional encodings exist in more complex networks and tasks as well as revealing an interesting future direction for adversarial attacks and defences based on position information.

\begin{figure}
\centering
  \begin{center}
  \resizebox{0.41\textwidth}{!}{ 
      \includegraphics[width=0.98\textwidth]{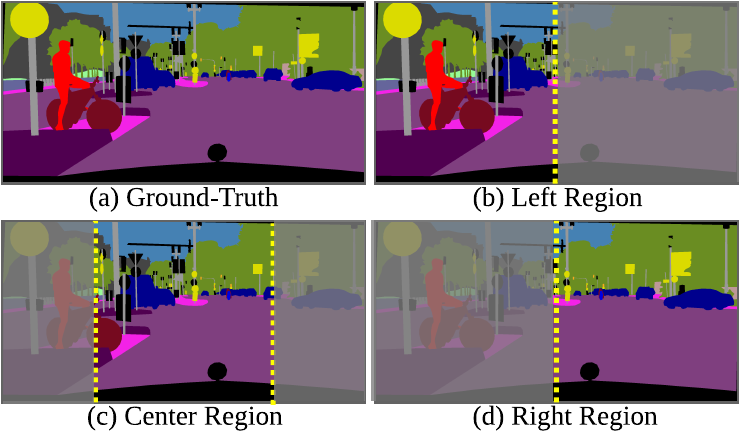}
      }
  \end{center}
  \vspace{-0.35cm}
  \caption{\textcolor{black}{Ground-truth samples with different evaluation regions (not highlighted) used in our experiments to demonstrate the degree of performance drop in specific image regions.}}
  \label{fig:location_gt}
  \vspace{-0.2cm}
\end{figure}


\subsubsection{Identifying Position-Specific Neurons} \label{sec:position_estimation}
\noindent \textbf{Overall Position-Encoding Channels.} Our first goal is to identify the channels in the latent representation of a network which encode the overall position of objects within an image. A simple and intuitive way to estimate the neurons which encode overall position information is by computing the \textit{absolute difference} between the activations of two latent representations of horizontally flipped image pairs. The key intuition of our approach is, given a network which encodes two images where the semantics are identical but the object's position is different, the channels which have small changes in activation are invariant to position, while those that have large changes in activation encode positional information. More formally, given a pretrained encoder for a pixel-wise task (e.g., semantic segmentation), $f(I)=z$, where $z \in \mathbb{R}^{1 \times 1 \times C}$ is a latent representation after passing it through a GAP layer, our goal is to estimate and rank the most position-specific neurons in $z$. For a given image, $I^a$, we simply apply horizontal flipping on $I^a$ to obtain the flipped image, $I^b$. Note that the only semantic factor that differs between this image pair is the \textit{absolute position} of the objects. We then feed these images to a pretrained network to obtain the latent representations, $\{z^a,z^b\}$. Finally, we compute the \textit{absolute} difference, $|\Delta z_i| = |z_i^a - z_i^b|$ between these two latent representations, where $i$ denotes the sample index from dataset. We calculate $|\Delta z_i|$ for all images, $i$, in the Cityscapes~\cite{cordts2016cityscapes} validation set and average the differences to obtain a position encoding score for each neuron. Finally, we rank the neurons in descending order to obtain an ordered list of the overall position-specific neurons, $\hat{z}\in \mathbb{R}^{1 \times 1 \times C}$, where the first element has the highest amount of \textit{positional encoding} and the final element has the largest \textit{positional invariance}. We can formalize $\hat{z}$ as follows:
\begin{equation}\label{eq:rank_pos_abs}
    \hat{z} = \texttt{argsort}_{j \in C} \Big [ \frac{1}{|\mathcal{D}|} \sum_{i=1}^{|\mathcal{D}|} |\Delta z_i| \Big] \ .
\end{equation}

\begin{figure*}
	\begin{center}
     \centering 
		\resizebox{0.98\textwidth}{!}{
\begin{tikzpicture} \ref{target_legend}
    \begin{axis}[
       line width=1.0,
        title={ (a) Overall mIoU},
        title style={at={(axis description cs:0.5,0.95)},anchor=north,font=\normalsize},
        xlabel={Top $N$ Neurons Removed},
        xmin=0, xmax=550,
        ymin=63, ymax=76,
        xtick={50,100,200,300,400,500},
        ytick={64,66,68,70,72,74,76},
        x tick label style={font=\footnotesize},
        y tick label style={font=\footnotesize},
        x label style={at={(axis description cs:0.5,0.03)},anchor=north,font=\small},
        y label style={at={(axis description cs:0.12,.5)},anchor=south,font=\normalsize},
        width=6.7cm,
        height=5.5cm,        
        ymajorgrids=false,
        xmajorgrids=true,
        major grid style={dotted,green!20!black},
        legend style={
         draw=none,
         nodes={scale=0.85, transform shape},
         cells={anchor=west},
         legend style={at={(1.17,0.25)},anchor=south}, font =\footnotesize},
         legend entries={[black]Normal,[black]Kernel-Flip*,[black]Kernel-Flip**,[black]Random,[black]Baseline},
        legend to name=target_legend,
    ]
    \addplot[line width=1.8pt,mark size=1.9pt,color=orange,mark=*,]
        coordinates {(50,72.8)(100,71.9)(200,70.2)(300,69.6)(400,67.9)(500,64.7)};
        
    \addplot[line width=1.8pt,mark size=1.9pt,color=cyan,mark=triangle,]
        coordinates {(50,72.9)(100,72.2)(200,71.2)(300,69.1)(400,67.9) (500,66.7)};
        
    \addplot[line width=1.8pt,mark size=1.9pt,color=purple,mark=diamond*,]
        coordinates {(50,72.7)(100,72.3)(200,71.4)(300,69.8)(400,67.6) (500,65.6)};

    
    \addplot[line width=1.8pt,mark size=1.9pt,color=red,mark=square*,]
        coordinates {(50,73.8)(100,73.5)(200,72.9) (300, 72.7) (400,71.6) (500,70.5)};
        
    \addplot[line width=1.8pt, black,dotted,sharp plot,update limits=false] 
	    coordinates {(0,74)(550,74)};
    \end{axis}
\end{tikzpicture}
\begin{tikzpicture} 
    \begin{axis}[
       line width=1.0,
        title={(b) Left Half Region},
        title style={at={(axis description cs:0.5,0.95)},anchor=north,font=\normalsize},
        xlabel={Top $N$ Neurons Removed},
        xmin=0, xmax=550,
        ymin=63, ymax=76,
        xtick={50,100,200,300,400,500},
        ytick={64,66,68,70,72,74,76},
        x tick label style={font=\footnotesize},
        y tick label style={font=\footnotesize},
        x label style={at={(axis description cs:0.5,0.03)},anchor=north,font=\small},
        y label style={at={(axis description cs:0.12,.5)},anchor=south,font=\normalsize},
        width=6.7cm,
        height=5.5cm,        
        ymajorgrids=false,
        xmajorgrids=true,
        major grid style={dotted,green!20!black},
        legend style={
         draw=none,
         nodes={scale=0.85, transform shape},
         cells={anchor=west},
         legend style={at={(1.17,0.25)},anchor=south}, font =\footnotesize},
         legend entries={[black]Normal,[black]Kernel-Flip*,[black]Kernel-Flip**,[black]Random,[black]Baseline},
        legend to name=target_legend,
    ]
        
    \addplot[line width=1.8pt,mark size=1.9pt,smooth,color=orange,mark=square*,]
        coordinates {(50, 71.5)(100,70.5)(200,68.5)(300,68.2) (400, 66.6) (500, 63.7)};
        
    \addplot[line width=1.8pt,mark size=1.9pt,color=cyan,mark=triangle,]
        coordinates {(50,71.6)(100,70.9)(200,69.8)(300,67.6)(400,66.2) (500,65.0)};
        
    \addplot[line width=1.8pt,mark size=1.9pt,color=purple,mark=diamond*,]
        coordinates {(50,71.3)(100,71.1)(200,70.2)(300,68.7)(400,66.9) (500,65.0)};
        
    \addplot[line width=1.8pt,mark size=1.9pt,color=red,mark=*,]
        coordinates {(50,72.8)(100,72.4)(200,71.0)(300,70.5) (400, 70.4) (500, 69.3)};
    \addplot[line width=1.8pt, black,dotted,sharp plot,update limits=false] 
	    coordinates {(0,72.8)(550,72.8)};
    \end{axis}
\end{tikzpicture}

\begin{tikzpicture} 
    \begin{axis}[
       line width=1.0,
        title={(c) Right Half Region},
        title style={at={(axis description cs:0.5,0.95)},anchor=north,font=\normalsize},
        xlabel={Top $N$ Neurons Removed},
        xmin=0, xmax=550,
        ymin=63, ymax=76,
        xtick={50,100,200,300,400,500},
        ytick={64,66,68,70,72,74,76},
        x tick label style={font=\footnotesize},
        y tick label style={font=\footnotesize},
        x label style={at={(axis description cs:0.5,0.03)},anchor=north,font=\small},
        y label style={at={(axis description cs:0.12,.5)},anchor=south,font=\normalsize},
        width=6.7cm,
        height=5.5cm,        
        ymajorgrids=false,
        xmajorgrids=true,
        major grid style={dotted,green!20!black},
        legend style={
         draw=none,
         nodes={scale=0.85, transform shape},
         cells={anchor=west},
         legend style={at={(1.17,0.25)},anchor=south}, font =\footnotesize},
         legend entries={[black]Normal,[black]Kernel-Flip*,[black]Kernel-Flip**,[black]Random,[black]Baseline},
        legend to name=target_legend,
    ]
        
    \addplot[line width=1.8pt,mark size=1.9pt,smooth,color=orange,mark=square*,]
        coordinates {(50,72.9)(100,72.5)(200,71.6)(300,70.5) (400, 68.7) (500, 65.1)};
        
    \addplot[line width=1.8pt,mark size=1.9pt,color=cyan,mark=triangle,]
        coordinates {(50,72.8)(100,72.5)(200,71.8)(300,70.0)(400,68.7) (500,67.5)};
        
    \addplot[line width=1.8pt,mark size=1.9pt,color=purple,mark=diamond*,]
        coordinates {(50,72.8)(100,72.1)(200,71.1)(300,69.2)(400,66.6) (500,64.5)};

    \addplot[line width=1.8pt,mark size=1.9pt,color=red,mark=*,]
        coordinates {(50, 73.5)(100,73.2)(200,72.9)(300,72.6) (400, 71.7) (500, 70.9)};
    \addplot[line width=1.8pt, black,dotted,sharp plot,update limits=false] 
	    coordinates {(0,73.8)(550,73.8)};
    \end{axis}
\end{tikzpicture}

\begin{tikzpicture} 
    \begin{axis}[
       line width=1.0,
        title={(d) Center Region},
        title style={at={(axis description cs:0.5,0.95)},anchor=north,font=\normalsize},
        xlabel={Top $N$ Neurons Removed},
        xmin=0, xmax=550,
        ymin=63, ymax=76,
        xtick={50,100,200,300,400,500},
        ytick={64,66,68,70,72,74,76},
        x tick label style={font=\footnotesize},
        y tick label style={font=\footnotesize},
        x label style={at={(axis description cs:0.5,0.03)},anchor=north,font=\small},
        y label style={at={(axis description cs:0.12,.5)},anchor=south,font=\normalsize},
        width=6.7cm,
        height=5.5cm,        
        ymajorgrids=false,
        xmajorgrids=true,
        major grid style={dotted,green!20!black},
        legend style={
         draw=none,
         nodes={scale=1.0, transform shape},
         cells={anchor=west},
         legend style={at={(1.5,0.11)},anchor=south}, font =\footnotesize},
         legend entries={[black]Normal,[black]Kernel-Flip*,[black]Kernel-Flip**,[black]Random,[black]Baseline},
        legend to name=target_legend,
    ]
        
    \addplot[line width=1.8pt,mark size=1.9pt,smooth,color=orange,mark=square*,]
        coordinates {(50,70.8)(100,70.2)(200,68.8)(300,68.4) (400, 67.1) (500, 64.3)};
        
    \addplot[line width=1.8pt,mark size=1.9pt,color=cyan,mark=triangle,]
        coordinates {(50,70.9)(100,70.4)(200,69.7)(300,68.2)(400,66.9) (500,66.0)};
        
    \addplot[line width=1.8pt,mark size=1.9pt,color=purple,mark=diamond*,]
        coordinates {(50,70.5)(100,70.1)(200,69.1)(300,67.4)(400,65.5) (500,63.8)};

    \addplot[line width=1.8pt,mark size=1.9pt,color=red,mark=*,]
        coordinates {(50,71.7)(100,71.4)(200,70)(300,69.6) (400, 69.4) (500, 68.6)};
    \addplot[line width=1.8pt, black,dotted,sharp plot,update limits=false] 
	    coordinates {(0,71.9)(550,71.9)};
    \end{axis}
\end{tikzpicture}

  }
  \hfill
  \resizebox{0.98\textwidth}{!}{

  \begin{tikzpicture} \ref{target_legend_3}
    \begin{axis}[
       line width=1.0,
        title={Overall mIoU},
        title style={at={(axis description cs:0.5,0.95)},anchor=north,font=\normalsize},
        xlabel={Ranked Neurons Removed},
        xmin=-50, xmax=450,
        ymin=71.5, ymax=75,
        xtick={0,100,200,300,400},
        xticklabels={0-100,100-200,200-300,300-400,400-500},
        ytick={72,73,74},
        x tick label style={font=\footnotesize,anchor=north},
        y tick label style={font=\footnotesize},
        x label style={at={(axis description cs:0.5,0.02)},anchor=north,font=\small},
        y label style={at={(axis description cs:0.12,.5)},anchor=south,font=\normalsize},
        width=6.9cm,
        height=5.5cm,        
        ymajorgrids=false,
        xmajorgrids=true,
        major grid style={dotted,green!20!black},
        legend style={
         draw=none,
         nodes={scale=0.85, transform shape},
         cells={anchor=west},
         legend style={at={(3.95,0.25)},anchor=south}, font =\small},
         legend entries={[black]Normal,[black]Random,[black]Baseline},
        legend to name=target_legend_3,
    ]
    \addplot[line width=1.8pt,mark size=1.9pt,color=orange,mark=*,]
        coordinates {(0,71.9)(100,72.97)(200,73.4)(300,73.2)(400,73.5)};
        
     \addplot[line width=1.8pt,mark size=1.9pt,color=cyan,mark=*,]
         coordinates {(0,72.2)(100,73)(200,73.1)(300,73.4)(400,73.6)};
         
     \addplot[line width=1.8pt,mark size=1.9pt,color=purple,mark=*,]
         coordinates {(0,72.3)(100,73.3)(200,73.2)(300,73.2)(400,73.7)};
    
    \addplot[line width=1.8pt,mark size=1.9pt,color=red,mark=square*,]
        coordinates {(0,73.5)(100,73.5)(200,73.5)(300,73.5)(400,73.5)};
        
    \addplot[line width=1.8pt, black,dotted,sharp plot,update limits=false] 
	    coordinates {(-50,74)(450,74)};
    \end{axis}
\end{tikzpicture}

\begin{tikzpicture}
    \begin{axis}[
       line width=1.0,
        title={Left Half Region},
        title style={at={(axis description cs:0.5,0.95)},anchor=north,font=\normalsize},
        xlabel={Ranked Neurons Removed},
        xmin=-50, xmax=450,
        ymin=70, ymax=74,
        xtick={0,100,200,300,400,500},
        xticklabels={0-100,100-200,200-300,300-400,400-500},
        ytick={70,71,72, 73},
        x tick label style={font=\footnotesize,anchor=north},
        y tick label style={font=\footnotesize},
        x label style={at={(axis description cs:0.5,0.02)},anchor=north,font=\small},
        y label style={at={(axis description cs:0.12,.5)},anchor=south,font=\normalsize},
        width=6.9cm,
        height=5.5cm,        
        ymajorgrids=false,
        xmajorgrids=true,
        major grid style={dotted,green!20!black},
        legend style={
         draw=none,
         nodes={scale=0.85, transform shape},
         cells={anchor=west},
         legend style={at={(3.95,0.25)},anchor=south}, font =\small},
         legend entries={[black]Normal,[black]Random,[black]Baseline},
        legend to name=target_legend_3,
    ]
    \addplot[line width=1.8pt,mark size=1.9pt,color=orange,mark=*,]
        coordinates {(0,70.6)(100,71.4)(200,72.5)(300,71.9)(400,72.2)};
        
     \addplot[line width=1.8pt,mark size=1.9pt,color=cyan,mark=*,]
         coordinates {(0,70.8)(100,71.5)(200,71.7)(300,72.1)(400,72.4)};
         
     \addplot[line width=1.8pt,mark size=1.9pt,color=purple,mark=*,]
         coordinates {(0,71.1)(100,72.1)(200,71.988)(300,72.1)(400,72.6)};
    
    \addplot[line width=1.8pt,mark size=1.9pt,color=red,mark=square*,]
        coordinates {(0,72.4)(100,72.4)(200,72.4)(300,72.4)(400,72.4)};
        
    \addplot[line width=1.8pt, black,dotted,sharp plot,update limits=false] 
	    coordinates {(-50,72.8)(450,72.8)};
    \end{axis}
\end{tikzpicture}

\begin{tikzpicture}
    \begin{axis}[
       line width=1.0,
        title={Right Half Region},
        title style={at={(axis description cs:0.5,0.95)},anchor=north,font=\normalsize},
        xlabel={Ranked Neurons Removed},
        xmin=-50, xmax=450,
        ymin=72, ymax=74.5,
        xtick={0,100,200,300,400,500},
        xticklabels={0-100,100-200,200-300,300-400,400-500},
        ytick={72, 73,74},
        x tick label style={font=\footnotesize,anchor=north},
        y tick label style={font=\footnotesize},
        x label style={at={(axis description cs:0.5,0.02)},anchor=north,font=\small},
        y label style={at={(axis description cs:0.12,.5)},anchor=south,font=\normalsize},
        width=6.9cm,
        height=5.5cm,        
        ymajorgrids=false,
        xmajorgrids=true,
        major grid style={dotted,green!20!black},
        legend style={
         draw=none,
         nodes={scale=0.85, transform shape},
         cells={anchor=west},
         legend style={at={(3.95,0.25)},anchor=south}, font =\small},
         legend entries={[black]Normal,[black]Random,[black]Baseline},
        legend to name=target_legend_3,
    ]
    \addplot[line width=1.8pt,mark size=1.9pt,color=orange,mark=*,]
        coordinates {(0,72.2)(100,72.5)(200,72.8)(300,73.4)(400,73.5)};
        
     \addplot[line width=1.8pt,mark size=1.9pt,color=cyan,mark=*,]
         coordinates {(0,72.5)(100,72.8)(200,72.95)(300,72.8)(400,73.35)};
         
     \addplot[line width=1.8pt,mark size=1.9pt,color=purple,mark=*,]
         coordinates {(0,72.1)(100,72.8)(200,72.95)(300,72.8)(400,73.3)};
    
    \addplot[line width=1.8pt,mark size=1.9pt,color=red,mark=square*,]
        coordinates {(0,73.2)(100,73.2)(200,73.2)(300,73.2)(400,73.2)};
        
    \addplot[line width=1.8pt, black,dotted,sharp plot,update limits=false] 
	    coordinates {(-50,73.8)(450,73.8)};
    \end{axis}
\end{tikzpicture}

\begin{tikzpicture}
    \begin{axis}[
       line width=1.0,
        title={Left Half Region},
        title style={at={(axis description cs:0.5,0.95)},anchor=north,font=\normalsize},
        xlabel={Center Region},
        xmin=-50, xmax=450,
        ymin=69.8, ymax=72.5,
        xtick={0,100,200,300,400,500},
        xticklabels={0-100,100-200,200-300,300-400,400-500},
        ytick={70,71,72},
        x tick label style={font=\footnotesize,anchor=north},
        y tick label style={font=\footnotesize},
        x label style={at={(axis description cs:0.5,0.02)},anchor=north,font=\small},
        y label style={at={(axis description cs:0.12,.5)},anchor=south,font=\normalsize},
        width=6.9cm,
        height=5.5cm,        
        ymajorgrids=false,
        xmajorgrids=true,
        major grid style={dotted,green!20!black},
        legend style={
         draw=none,
         nodes={scale=0.85, transform shape},
         cells={anchor=west},
         legend style={at={(3.95,0.04)},anchor=south}, font =\small},
         legend entries={[black]Normal,[black]Kernel-Flip*, [black]Kernel-Flip**,[black]Random,[black]Baseline},
        legend to name=target_legend_3,
    ]
    \addplot[line width=1.8pt,mark size=1.9pt,color=orange,mark=*,]
        coordinates {(0,70.2)(100,70.989)(200,71.3567)(300,71.3)(400,71.329)};
        
     \addplot[line width=1.8pt,mark size=1.9pt,color=cyan,mark=*,]
         coordinates {(0,70.4)(100,71.2)(200,71.29)(300,71.32)(400,71.6)};
         
     \addplot[line width=1.8pt,mark size=1.9pt,color=purple,mark=*,]
         coordinates {(0,70.0)(100,71.0)(200,71.2)(300,71.2)(400,71.6)};
    
    \addplot[line width=1.8pt,mark size=1.9pt,color=red,mark=square*,]
        coordinates {(0,71.4)(100,71.4)(200,71.4)(300,71.4)(400,71.4)};
        
    \addplot[line width=1.8pt, black,dotted,sharp plot,update limits=false] 
	    coordinates {(-50,71.9)(450,71.9)};
    \end{axis}
\end{tikzpicture}
}
	\end{center}
	\vspace{-0.35cm}
	\caption{Semantic segmentation results for DeepLabv3-ResNet-50 trained on Cityscapes. \textbf{Top Row:}  The top $N$ overall \textit{position}-specific channels are removed from the latent representation during inference. Removing the top $N$ position-specific channels from the \textit{DeepLabv3-ResNet-50} impairs the network's recognition ability more severely compared to the removal of the same number of \textit{random} channels. \blue{\textbf{Bottom Row:} Same settings as the top row, but only channels in the ranges specified (i.e., every 100 channels) are removed.}}\label{fig:remove_neuron_target}
	\vspace{-0.3cm}
\end{figure*}
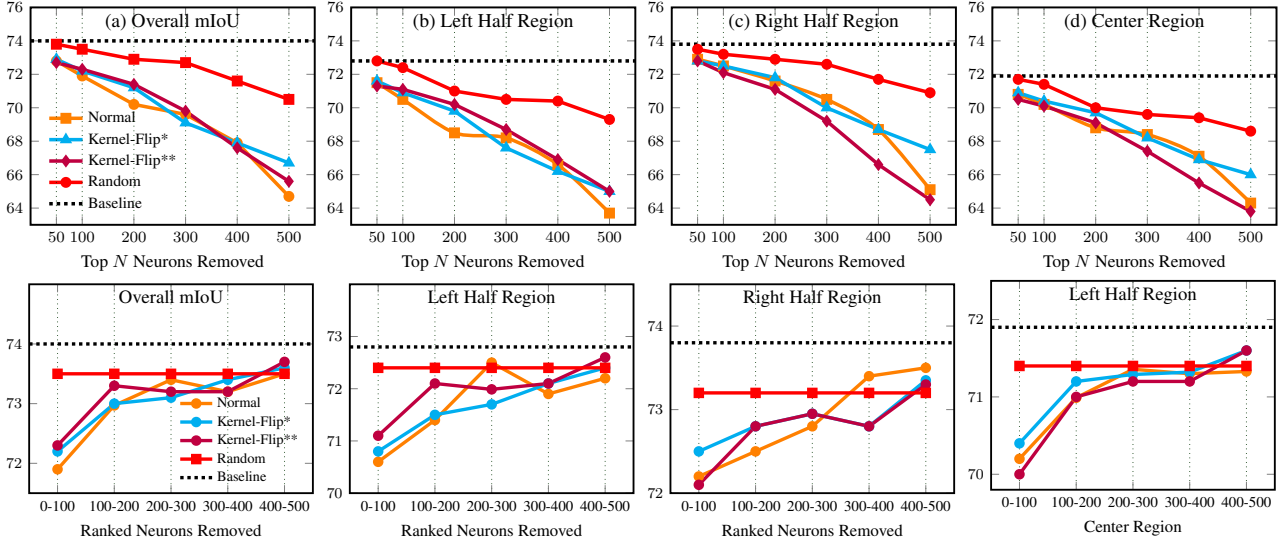

\blue{One might argue that a horizontally flipped input will introduce differences to the network even without considering the positional information because convolutional kernels can be asymmetric~\cite{alsallakh2020mind}. To further justify our ranking strategy we also rank overall position encoding channels under two settings. Kernel-Flip$^\ast$: We pass an image and its flipped version (Eq.~\ref{eq:rank_pos_abs}) to a pretrained model that has flipped kernels, Kernel-Flip$^{\ast\ast}$: We pass the original image to two pretrained models, with and without flipped kernels, and use the activation differences to rank the neurons using Eq.~\ref{eq:rank_pos_abs}.} \\

\noindent \textbf{Region-Specific Channels.} Contrasting Eq.~\ref{eq:rank_pos_abs} which treats any change in a channel's activation subject to horizontally flipping the image as an indication it encodes positional information, we now aim to identify and rank the channels based on encoding position for a \textit{specific} image region. Thus, we are interested in channels which are \textit{highly activated when an object is in a particular location, and have a low activation otherwise}. To this end, we create two subsets of data out of the Cityperson~\cite{cordts2016cityscapes} validation set: images where the pedestrians are only located on the left half of the image (val-left), and pedestrians that are only located in the right half (val-right (see Fig.~\ref{fig:location_gt}). If we want to identify the channels which encode objects on the left half of the image, we pass an image from \textit{val-left}, $I^a$, and its corresponding flipped image, $I^b$, through the network and take the \textit{signed difference} of the two latent representations after the GAP layer. After ranking the channels based on the average activation difference over the val-left dataset, we can now identify the channels which particularly activate for pedestrians on the left side. More specifically, we can calculate an ordered list of the left region-specific neurons, $\hat{z}^l\in \mathbb{R}^{1 \times 1 \times C}$ with the following equation: 

\begin{equation}\label{eq:rank_pos_sign}
    \hat{z}^l = \texttt{argsort}_{j \in C} \Big [ \frac{1}{|\mathcal{D}|} \sum_{i=1}^{|\mathcal{D}|} \Delta z_i \Big] \ .
\end{equation}

We can easily calculate an ordered list of the \textit{right} region-specific neurons $\hat{z}^r$ by simply passing val-right through the network and then follow the same procedure outlined above. Note that both of these procedures use the assumption that the ordering of the channels after a GAP layer admits positional information, both overall and region specific.

\subsubsection{Targeting Overall Position-Encoding Neurons} \label{sec:target_segmentation}
\noindent \textbf{Semantic Segmentation.} We first validate how the top $N$ overall position-encoding channels affect the performance~\cite{islam2020shape} of a state-of-the-art semantic segmentation network, DeepLabv3-ResNet-50~\cite{chen2017rethinking}, trained on the Cityscapes~\cite{cordts2016cityscapes} dataset. The goal of this evaluation is to determine how much DeepLabv3-ResNet-50 relies on these global position-encoding channels for semantic segmentation by measuring the difference in validation performance, measured by the mean intersection over union (mIoU), after \textit{removing} the top $N$ position-specific channels. We remove these $N$ neurons from the latent representation by simply setting the feature activations at these channel dimensions to \textit{zero}. In addition to the standard mIoU, we aim to evaluate whether these neurons affect all locations equally. To this end, we also assess the performance on three input regions: the \textit{left}, \textit{center} or \textit{right} region of the image. To perform this assessment, this we simply compute the mean intersection over union (mIoU) only on the \textit{left}, \textit{center}, or \textit{right} regions by setting the other pixels to the \textit{unlabelled value} (see Fig.~\ref{fig:location_gt} for an example). Note that each region (i.e., left, right, and center) has a spatial resolution of $1024\times1024$ pixels as the width of the Cityscapes images are 2048 pixels (note that the center overlaps with the left and right sides equally). In these experiments, we perform validation on the \textit{val} split of the Cityscapes dataset. \\ 

\noindent \textbf{Results.} Figure~\ref{fig:remove_neuron_target} presents the semantic segmentation results of \textit{DeepLabv3-ResNet-50} on Cityscapes in terms of mIoU when the top $N$ overall position-specific channels are set to \textit{zero}. Note that for this experiment, we do not fine-tune the pretrained segmentation network. Interestingly, we observe that gradually removing the position-specific neurons from Cityscapes pretrained DeepLabv3-ResNet-50 (with a baseline performance of 74.0\% mIoU) model hurts the overall mIoU (Fig.~\ref{fig:remove_neuron_target} (a)) significantly more than removing randomly selected neurons \blue{(note that `Normal' refers to neurons selected using Eq.~\ref{eq:rank_pos_abs} without flipping the kernels)}. For the position-specific neurons, removing the top 100 channels results in a performance of 71.9\%, while 100 random neurons drops the performance to only 73.4\%. When removing even more neurons, the difference is more significant. For example, removing 500 position-specific channels results in a performance of 64.7\% mIoU, a 9.3\% drop from the baseline of 74.0\%, while 500 random channels only drops the performance by 3.5\% to 70.5\% mIoU. \blue{When removing neurons using the kernel-flipping rankings a similar pattern in seen as removing these neurons degrades the performance more severely than removing random ones. }

\begin{figure*}
\centering
  \begin{center}
  \resizebox{0.98\textwidth}{!}{ 
      \includegraphics[width=0.99\textwidth]{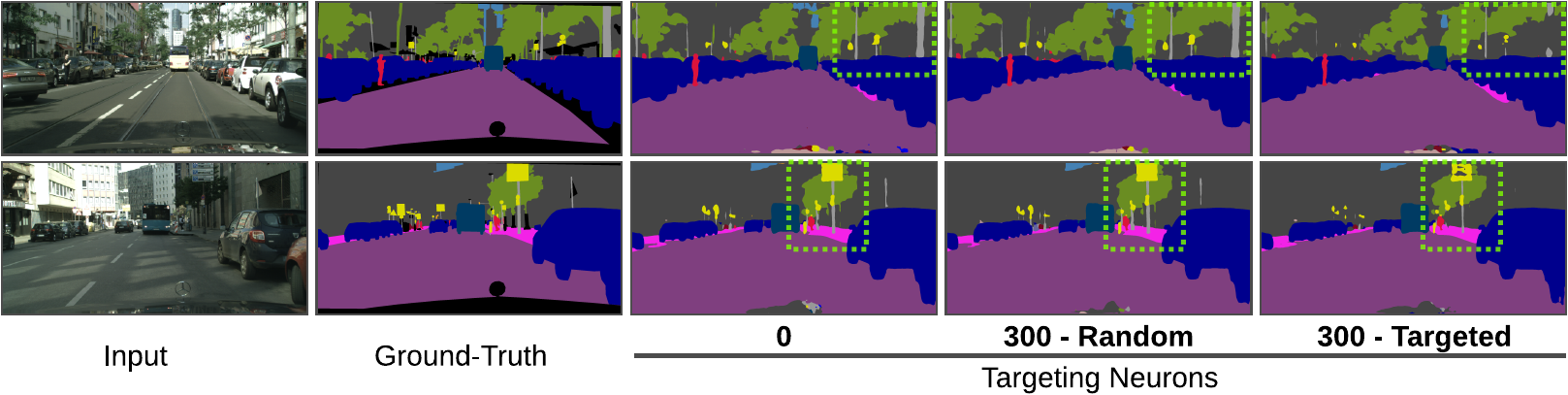}
      }
  \end{center}
  \vspace{-0.35cm}
  \caption{\textcolor{black}{Qualitative comparisons between the \textit{position-specific} and \textit{random} neuron removal on the Cityscapes val set. Note the performance drop on objects near the periphery (highlighted in dotted box) are particularly pronounced for our position specific neuron targeting.}}
  \label{fig:seg_prediction_examples}
\end{figure*}

These results clearly demonstrate the network's reliance on the channel-wise positional encoding in the latent representations to make accurate semantic segmentation predictions. 
Note that the performance drop by removing 100 position encoding neurons (a drop of 2.1\%) is quite significant as the size of the latent dimension, $z$, is 2048 for the \textit{DeepLabv3-ResNet-50} network. This finding is \textit{consistent for all locations} in the image. Figure~\ref{fig:remove_neuron_target} (b, c, d) shows the mIoU for the \textit{left}, \textit{right}, and \textit{center} regions (as seen in Fig.~\ref{fig:location_gt}). For each region, the performance drops more when targeting the position neurons obtained through the channel-wise ranking strategy than compared with targeting random neurons. \blue{Figure~\ref{fig:remove_neuron_target} (bottom row) plots the results for each portion of the image when we remove only 100 neurons in the ranking. These results also show that, in general, removing higher ranked neurons lowers the mIoU more than the ones at lower rankings.} Collectively, these results provide strong evidence that the ranked channels using Eq.~\ref{eq:rank_pos_abs} encode overall image position. Note that the center region has a lower baseline accuracy, and therefore a lower drop in performance, due to pixels in the center region of autonomous driving datasets being associated with objects which are a greater distance from the camera and therefore harder to accurately classify. 

Figure~\ref{fig:seg_prediction_examples} provides qualitative results on the Cityscapes validation images while removing $N$ specific neurons. It is clear that the segmentation quality gradually degrades with the increase of $N$. Note that the failure of segmenting the \textit{smaller} and \textit{thinner} objects in the left or right \textit{periphery} are particularly pronounced in our neuron specific targeting. \blue{We also validate how the top $N$ overall position-encoding channels affect the performance of a pedestrian detection network and report the results in the supplementary.} \\

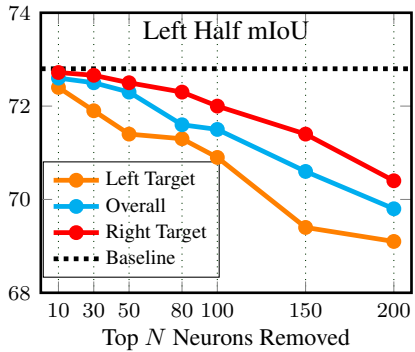
\begin{figure}
	\begin{center}
     \centering 
		\resizebox{0.33\textwidth}{!}{

\begin{tikzpicture} \ref{target_legend_2}
    \begin{axis}[
       line width=1.0,
        title={Left Half mIoU},
        title style={at={(axis description cs:0.5,0.95)},anchor=north,font=\normalsize},
        xlabel={Top $N$ Neurons Removed},
        xmin=0, xmax=210,
        ymin=68, ymax=74,
        xtick={10,30, 50, 80, 100, 150, 200},
        ytick={68,70,72,74},
        x tick label style={font=\footnotesize},
        y tick label style={font=\footnotesize},
        x label style={at={(axis description cs:0.5,0.05)},anchor=north,font=\small},
        y label style={at={(axis description cs:0.12,.5)},anchor=south,font=\normalsize},
        width=6.5cm,
        height=5.3cm,        
        ymajorgrids=false,
        xmajorgrids=true,
        major grid style={dotted,green!20!black},
        legend style={
         nodes={scale=0.85, transform shape},
         cells={anchor=west},
         legend style={at={(1.2,0.2)},anchor=south}, font =\small},
         legend entries={[black]Left Target,[black]Overall,[black]Right Target,[black]Baseline},
        legend to name=target_legend_2,
    ]
    \addplot[line width=1.8pt,mark size=1.9pt,color=orange,mark=*,]
        coordinates {(10,72.4)(30,71.9)(50,71.4) (80, 71.3) (100, 70.9) (150, 69.4) (200, 69.1)};
        
    \addplot[line width=1.8pt,mark size=1.9pt,color=cyan,mark=*,]
        coordinates {(10,72.6)(30,72.5)(50,72.3) (80, 71.6) (100, 71.5) (150, 70.6) (200, 69.8)};
        
    \addplot[line width=1.8pt,mark size=1.9pt,color=red,mark=*,]
        coordinates {(10,72.72)(30,72.66)(50,72.5) (80, 72.3) (100, 72.0) (150, 71.4) (200, 70.4)};
        
        
    \addplot[line width=1.8pt, black,dotted,sharp plot,update limits=false] 
	    coordinates {(0,72.8)(510,72.8)};
    \end{axis}
\end{tikzpicture}
  }
	\end{center}
	\vspace{-0.35cm}
	\caption{Semantic segmentation results (mIoU) evaluated on only the \textit{left half of the image} for DeepLabv3's trained on Cityscapes when the top $N$ \textit{region}-specific channels are removed from the latent representation during inference. Removing the top $N$ left-specific channels from the \textit{DeepLabv3-ResNet-50} impairs the network's recognition ability more severely compared to the corresponding removal of overall position or right-specific channels.}\label{fig:ss_loc_specific}
	\vspace{-0.2cm}
\end{figure}
	
\vspace{-0.5cm}
\subsubsection{Targeting Region-Specific Channels}
\noindent \textbf{Semantic Segmentation.} We now provide evidence that it is possible to harm the performance in a specific input region, for a fully convolutional neural network. We choose again the DeepLabv3-ResNet-50~\cite{chen2017rethinking} semantic segmentation network, trained on the Cityscapes~\cite{cordts2016cityscapes} dataset. We use the \textit{signed} ranking from Eq.~\ref{eq:rank_pos_sign} along with the \textit{val-left} and \textit{val-right} subsets to obtain the channel ranking in terms of region specific positional encoding. As motivation, we take the role of an adversary intent on causing a high-speed collision. As the Cityscapes~\cite{cordts2016cityscapes} dataset was collected in Germany, cars drive on the right side of the road. Therefore \textit{oncoming traffic appears on the left side of the images.} \blue{Our main motivation is to \textit{expose the possibility} of a more fine-grained region-specific attack for which we use a driving dataset as a test-case. There may be cases where adversaries want to impair performance in regions of interest, while minimizing the chance of being exposed.}
We target the semantic segmentation performance in the \textit{left-half of the image} using our channel targeting technique described in Sec.~\ref{sec:position_estimation}.

Figure~\ref{fig:ss_loc_specific} shows the validation results in terms of mIoU where only the left half of the image is considered during evaluation (see Fig.~\ref{fig:location_gt} for example ground-truth segmentation maps) and the top $N$ channels are set to \textit{zero}. Our hypothesis predicts that the left target curve should be lower than the right target curve as the left encoding channels should harm the mIoU on the left half more than the right encoding channels. As expected, the channels which we identified as encoding `Left' are predictive of a larger performance drop in the left-half mIoU calculation compared with the other channels. Interestingly, the `Overall' position channels have the next largest performance drop, followed by the `Right' channels. This provides further evidence that these channels do capture objects in specific regions in the image, as one would expect less overlap between the `Left' and `Right' channels compared with the `Left' and `Overall' position channels. Also note that (although small) a discrepancy in mIoU between the Left and Right channels is even observed after 10 neurons are zeroed out with performance drops of 0.6\% and 0.08\%, respectively. This discrepancy grows as more of the channels that capture objects on the left are removed. 
Results for an equivalent attack on the right half of the image are included in the supplementary and are consistent with findings presented here on the left half.

\section{Conclusion}
We have shown for the first time how CNNs with global average pooling layers, which collapse the spatial dimensions, admit absolute positional information. Moreover, we showed that position information is encoded based on the \textit{ordering} of the channels while semantic information is largely not. 
We then applied these findings to various real-world applications. We proposed an objective function to improve the translation invariance of CNNs trained for object recognition. 
We introduced a simple and intuitive technique to identify and rank the position-encoding neurons in a CNN's latent representation. We showed this technique could identify the channels which largely encode either (i) global position or (ii) region-specific positions in the input. Regarding the global position channels, we show that suppressing their responses in networks trained for semantic segmentation and object recognition results in a greater drop in performance compared to other baselines, suggesting that these CNNs rely significantly on the channel-wise positional encoding. Finally, we demonstrated the possibility of a fine-grained adversarial attack which aims to harm the performance of a network in a \textit{specific location}. All of these experiments were performed through the manipulation of the latent representations \textit{after a global average pooling layer} demonstrating the rich positional information contained in the ordering of the channels in a number of neural network architectures. We believe these findings and associated applications can help guide the future design of neural networks to contain the right inductive biases for the task at hand.

\vspace{0.1cm}

\noindent \textbf{Acknowledgements.}
We gratefully acknowledge financial support from the Canadian NSERC Discovery Grants and Vector Institute Post-graduate Affiliation award.  K.G.D. contributed to this work in his personal capacity as an Associate Professor at York University. We thank the NVIDIA Corporation for providing GPUs through their academic program.


{\small
\bibliographystyle{ieee_fullname}
\bibliography{main}
}




\newcommand{\hbAppendixPrefix}{S}
\renewcommand{\thefigure}{\hbAppendixPrefix\arabic{figure}}
\setcounter{figure}{0}
\renewcommand{\thetable}{\hbAppendixPrefix\arabic{table}}
\setcounter{table}{0}
\renewcommand{\theequation}{\hbAppendixPrefix\arabic{equation}}
\setcounter{equation}{0}
\renewcommand{\thesection}{\hbAppendixPrefix\arabic{section}}
\setcounter{section}{0}



\section{Decoding Absolute Location From Pre-trained Models}\label{sec:decode_pre}

We have shown in Sec.~\ref{sec:channel_position} that Global Average Pooling (GAP) layers can admit absolute position information by means of the \textit{ordering} of the channel dimensions. Now we explore how much absolute position information can be decoded from various \textit{pre-trained} models which are not explicitly trained for location classification. We first explore an ImageNet~\cite{ImageNet} pretrained ResNet-18 model~\cite{he2016deep}, $f_{\text{enc}}$. As input, we use the same images as described in Sec. 3 of the main manuscript: we place a CIFAR-10~\cite{krizhevsky2014cifar} image on a black canvas in a location (note there is no overlapping with other locations), where each location has a unique index (see Fig. 1 in the main manuscript for a visual example). We feed this grid-based input image, $I$, to $f_{\text{enc}}$ and obtain the latent representation, $z$. Next, we apply a $1\times1$ convolution on $z$ to produce a representation, $z^\prime$, which has the same number of channel dimensions as the number of classification logits. Then we apply the GAP operation which collapses the spatial dimension, resulting in the final classification logits, $\hat{y}$. Note that we freeze the classification network as we are interested in validating how much absolute location can be decoded from the latent representation of pre-trained model for image classification. We can formalize the operations as follows:
\begin{equation}
    z = f_{\text{enc}} (I), \hspace{0.2cm} z^\prime = \text{Conv}_{1\times 1} (z), \hspace{0.2cm} \hat{y}=\text{GAP} (z^\prime).
\end{equation}

We also decode absolute position information from the latent representation of a ResNet-18 model trained for the task of \textit{semantic segmentation} on the PASCAL VOC 2012 dataset~\cite{PASCALVOC}. The same method is applied as above, using a simple $1\times1$ convolution on the latent representation $z$, followed by a GAP layer to output the number of location classes.

\begin{figure}
	\begin{center}
     \centering 
		\resizebox{0.5\textwidth}{!}{
\begin{tikzpicture} \ref{legend_GAPSHUFFLE}
    \begin{axis}[
       line width=1.0,
        title={Pre-Trained Positional Encoding},
        title style={at={(axis description cs:0.5,1.1)},anchor=north,font=\normalsize},
        xlabel={Gridsize},
        ylabel={Accuracy (\%)},
        xmin=2.5, xmax=7.5,
        ymin=-10, ymax=115,
        xtick={3,5,7},
        ytick={20,40,60,80,100},
        x tick label style={font=\footnotesize},
        y tick label style={font=\footnotesize},
        x label style={at={(axis description cs:0.5,0.05)},anchor=north,font=\small},
        y label style={at={(axis description cs:0.12,.5)},anchor=south,font=\normalsize},
        width=6.5cm,
        height=5.5cm,        
        ymajorgrids=false,
        xmajorgrids=true,
        major grid style={dotted,green!20!black},
        legend style={
         nodes={scale=0.85, transform shape},
         cells={anchor=west},
         legend style={at={(6.5,1.1)},anchor=south}, font =\small},
         legend entries={[black]GAPNet-Cls, [black]GAPNet-Seg, [black]ShuffleNet-Cls,[black]ShuffleNet-Seg,},
        legend to name=legend_GAPSHUFFLE,
    ]
    \addplot[line width=1.8pt,mark size=1pt,color=cyan,mark=*,]
        coordinates {(3,93.6)(5,86.3)(7,75.2)};
    \addplot[line width=1.8pt,mark size=1pt,color=cyan,mark=square*,]
        coordinates {(3,99.5)(5,99.7)(7,97.1)};
    \addplot[line width=1.8pt,mark size=1pt,color=orange,mark=*,]
        coordinates {(3,16.7)(5,5.9)(7,3.0)};
    \addplot[line width=1.8pt,mark size=1pt,color=orange,mark=square*,]
        coordinates {(3,29.2)(5,12.0)(7,7.2)};
    \end{axis}
\end{tikzpicture}
  }
	\end{center}
	\vspace{-0.35cm}
	\caption{Decoding absolute location from ImageNet~\cite{ImageNet} pretrained classification and PASCAL VOC 2012~\cite{PASCALVOC} pretrained segmentation models (ResNet-18~\cite{he2016deep} backbone) using \textit{GAPNet} and \textit{ShuffleNet}. Note that we \textit{freeze} the classification and segmentation pretrained models and only train the GAP or linear layer which predicts the output logits. It is clear that \textit{GAPNet} can decode  positional information from a model trained for classification or semantic segmentation, while \textit{ShuffleNet} fails to correctly decode locations. This demonstrates that positions are encoded channel-wise in the latent representation.}\label{fig:supp_shuffle}
\end{figure}
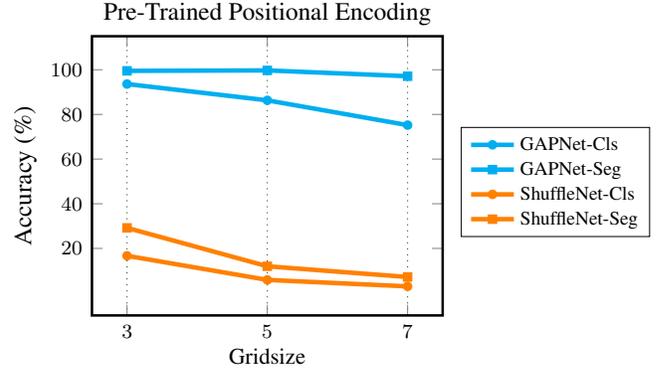

We provide the location classification results from image classification and semantic segmentation pretrained models in Fig.~\ref{fig:supp_shuffle}. These results are consistent with the results in Sec. 3 of the main manuscript and further demonstrate unequivocally that rich positional information is contained in the channels of CNNs. Furthermore, as shown by the degradation of performance when a shuffling operation is applied (\textit{ShuffleNet}), that this information is based on the \textit{ordering} of the channels.

\section{Shift Invariance Results}\label{sec:shift}

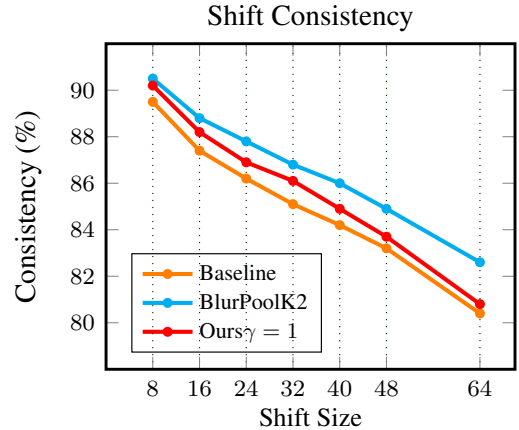
\begin{figure}
	\begin{center}
     \centering 
		\resizebox{0.4\textwidth}{!}{
\begin{tikzpicture} \ref{legend_shift}
    \begin{axis}[
       line width=1.0,
        title={Shift Consistency},
        title style={at={(axis description cs:0.5,1.1)},anchor=north,font=\normalsize},
        xlabel={Shift Size},
        ylabel={Consistency (\%)},
        xmin=0, xmax=70,
        ymin=78, ymax=92,
        xtick={8,16, 24, 32, 40, 48, 64},
        ytick={80,82,84,86,88,90},
        x tick label style={font=\footnotesize},
        y tick label style={font=\footnotesize},
        x label style={at={(axis description cs:0.5,0.05)},anchor=north,font=\small},
        y label style={at={(axis description cs:0.12,.5)},anchor=south,font=\normalsize},
        width=6.5cm,
        height=5.5cm,        
        ymajorgrids=false,
        xmajorgrids=true,
        major grid style={dotted,green!20!black},
        legend style={
         nodes={scale=0.85, transform shape},
         cells={anchor=west},
         legend style={at={(1.45,0.2)},anchor=south}, font =\small},
         legend entries={[black]Baseline, [black]BlurPoolK2, [black]Ours$\gamma=1$},
        legend to name=legend_shift,
    ]
    \addplot[line width=1.4pt,mark size=1pt,color=orange,mark=*,]
        coordinates {(8,89.5)(16,87.4)(24,86.2)(32,85.1)(40,84.2)(48,83.2)(64,80.4)};
        
    \addplot[line width=1.4pt,mark size=1pt,color=cyan,mark=*,]
        coordinates {(8,90.5)(16,88.8)(24,87.8)(32,86.8)(40,86)(48,84.9)(64,82.6)};
        
    \addplot[line width=1.4pt,mark size=1pt,color=red,mark=*,]
        coordinates {(8,90.2)(16,88.2)(24,86.9)(32,86.1)(40,84.9)(48,83.7)(64,80.8)};
        
        
    \end{axis}
\end{tikzpicture}
  }
	\end{center}
	\vspace{-0.35cm}
	\caption{Comparison of shifting consistency with increasing pixel shift sizes across different methods trained on ImageNet~\cite{ImageNet} .}\label{fig:supp_shift}
\end{figure}

In Table 2 of the main manuscript we presented shift consistency results for various networks. We show additional shift-consistency results in Figure~\ref{fig:supp_shift}. We compare three networks, a standard ResNet-50~\cite{he2016deep}, a ResNet-50 with BlurPool-k2~\cite{zhang2019making}, and our \textit{AugShift} method. Note that we train each model on ImageNet and use the validation set to validate the consistency for pixel $\text{shifts}= \{8,16,32,40,48,64\}$. Our method consistently outperforms the ResNet-50 baseline and reveals a useful adjunct strategy when compared with BlurPool.

\section{Targeting Region-Specific Channels}\label{sec:region_appen}

\begin{figure}
	\begin{center}
     \centering 
		\resizebox{0.42\textwidth}{!}{

\begin{tikzpicture} \ref{target_legend_7}
    \begin{axis}[
       line width=1.0,
        title={Right Channel Targeting},
        title style={at={(axis description cs:0.5,1.1)},anchor=north,font=\normalsize},
        xlabel={Top $N$ Neurons Removed},
        ylabel={Difference mIoU (\%)},
        xmin=-10, xmax=170,
        ymin=-2.5, ymax=0.5,
        xtick={0, 30, 50, 80, 100, 150},
        ytick={0, -1,-2,-3},
        x tick label style={font=\footnotesize},
        y tick label style={font=\footnotesize},
        x label style={at={(axis description cs:0.5,0.05)},anchor=north,font=\small},
        y label style={at={(axis description cs:0.16,.5)},anchor=south,font=\normalsize},
        width=6.5cm,
        height=5.5cm,        
        ymajorgrids=false,
        xmajorgrids=true,
        major grid style={dotted,green!20!black},
        legend style={
         nodes={scale=0.8, transform shape},
         cells={anchor=west},
         legend style={at={(1.7,0.2)},anchor=south}, font =\small},
         legend entries={[black]Left Region mIoU,[black]Right Region mIoU,[black]Baseline},
        legend to name=target_legend_7,
    ]
    \addplot[line width=1.4pt,mark size=1pt,color=blue,mark=*,]
        coordinates {(10,-0.1)(30,-0.1)(50,-0.3) (80, -0.5) (100, -0.8) (150, -1.4) };

    \addplot[line width=1.4pt,mark size=1pt,color=red,mark=*,]
        coordinates {(10,-0.1)(30,-0.4)(50,-0.5) (80, -0.9) (100, -1.1) (150, -1.8) };


                
        
        
    \addplot[line width=1.8pt, black,dotted,sharp plot,update limits=false] 
	    coordinates {(-10,0)(510,0)};
    \end{axis}
\end{tikzpicture}
  }
	\end{center}
	\vspace{-0.35cm}
	\caption{Relative performance drop in terms of mIoU when the top $N$ \textit{right-specific} neurons are removed from the left and right regions. Note that we evaluate on either the \textit{left half or right half of the image} for DeepLabv3-ResNet-50's~\cite{chen2017rethinking} trained on Cityscapes~\cite{cordts2016cityscapes} when the top $N$ \textit{region}-specific channels are removed from the latent representation during inference.}\label{fig:ss_loc_specific_1}
\end{figure}
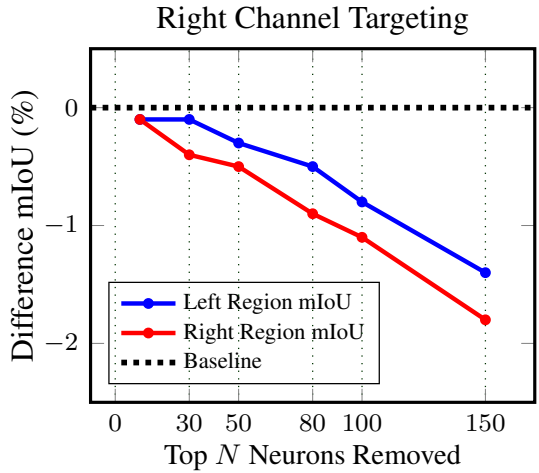

\begin{figure}
	\begin{center}
     \centering 
		\resizebox{0.42\textwidth}{!}{

\begin{tikzpicture} \ref{target_legend_9}
    \begin{axis}[
       line width=1.0,
        title={Left Channel Targeting},
        title style={at={(axis description cs:0.5,1.1)},anchor=north,font=\normalsize},
        xlabel={Top $N$ Neurons Removed},
        ylabel={Difference mIoU (\%)},
        xmin=-10, xmax=170,
        ymin=-4, ymax=0.5,
        xtick={0, 30, 50, 80, 100, 150},
        ytick={0, -1, -2, -3,-4},
        x tick label style={font=\footnotesize},
        y tick label style={font=\footnotesize},
        x label style={at={(axis description cs:0.5,0.05)},anchor=north,font=\small},
        y label style={at={(axis description cs:0.12,.5)},anchor=south,font=\normalsize},
        width=6.5cm,
        height=5.5cm,        
        ymajorgrids=false,
        xmajorgrids=true,
        major grid style={dotted,green!20!black},
        legend style={
         nodes={scale=0.85, transform shape},
         cells={anchor=west},
         legend style={at={(1.7,0.2)},anchor=south}, font =\small},
         legend entries={[black]Left Region mIoU,[black]Right Region mIoU,[black]Baseline},
        legend to name=target_legend_9,
    ]
    \addplot[line width=1.4pt,mark size=1pt,color=blue,mark=*,]
        coordinates {(10,-0.4)(30,-0.9)(50,-1.4) (80, -1.5) (100, -1.9) (150, -3.4)};
        
    \addplot[line width=1.4pt,mark size=1pt,color=red,mark=*,]
        coordinates {(10,-0)(30,-0.5)(50,-1.1) (80, -1.3) (100, -1.8) (150, -3.5) };
        
        
        
        
    \addplot[line width=1.8pt, black,dotted,sharp plot,update limits=false] 
	    coordinates {(-10,0)(510,0)};
    \end{axis}
\end{tikzpicture}
  }
	\end{center}
	\vspace{-0.35cm}
	\caption{Relative performance drop when the top $N$ \textit{left-specific} neurons are removed from the left and right regions. Note that we evaluate on either the \textit{left half or right half of the image} for DeepLabv3-ResNet-50's~\cite{chen2017rethinking} trained on Cityscapes~\cite{cordts2016cityscapes} when the top $N$ \textit{region}-specific channels are removed from the latent representation during inference. }\label{fig:ss_loc_specific_2}
\end{figure}
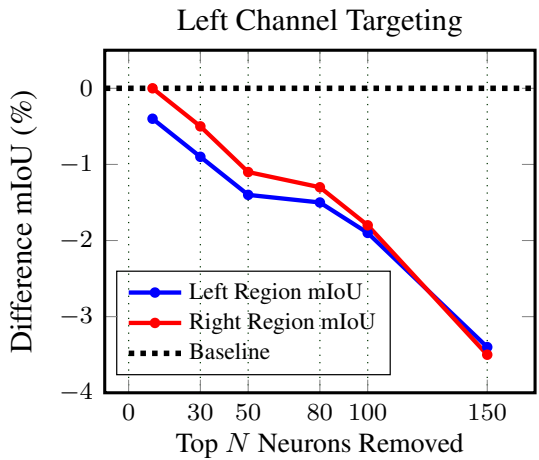

In Sec.~\ref{sec:target_segmentation} of the main manuscript, we have shown it is possible for specific channels in the latent representation of a CNN to encode specific regions contained in an image, and furthermore, that suppressing these activations can harm the performance in \textit{specific regions} in the image. We now show an overall comparison of the difference in performance, in terms of mean intersection over union (mIoU), between the left and right halves of the image, when either the left-encoding or right-encoding channels are suppressed. Figure~\ref{fig:ss_loc_specific_1} shows the change in mIoU when evaluated for the left and right halves of the Cityscapes~\cite{cordts2016cityscapes} validation image when the \textit{right}-encoding channels are turned off. As expected, we see a moderate but consistent decrease in performance on the \textit{right half} of the image. Figure~\ref{fig:ss_loc_specific_2} shows the same results but when targeting the \textit{left}-encoding channels. Similar to the right-encoding channels, we see a moderate but consistent decrease in performance on the \textit{left half} of the image. 

\begin{table}[t]
\def\arraystretch{1.07}
\setlength\tabcolsep{3.9pt}
\centering

\resizebox{0.49\textwidth}{!}{

	\begin{tabular}{l cccc c}

	
	\multirow{1}{*}{\rotatebox{0}{Methods}} && Reasonable $\downarrow$ & Small $\downarrow$ & Heavy$\downarrow$ & All$\downarrow$ \\
	 
	 \specialrule{1.2pt}{1pt}{1pt}
	  
	 Faster-RCNN~\cite{ren2015faster} && 10.3	&11.59	&33.07	&30.34 \\

	 + Random && 10.83	&11.87	&33.78	&31.92\\
	 
	 + \textbf{Targeted} && \textbf{12.51}&	\textbf{12.77}&	\textbf{36.95}	&\textbf{34.36} \\
	 
	 \midrule
	 
	 Cascade-RCNN~\cite{cai2018cascade} && 7.55&	8.55&	27.47&	26.89 \\

	 + Random && 7.85&	8.39&	28.31&	27.17\\
	 + \textbf{Targeted} && \textbf{8.44}	&\textbf{9.3}&\textbf{	30.59}&	\textbf{28.57} \\
	 
	 \midrule
	 
	 CSP~\cite{liu2019high} && 11.05 &14.76	&41.35&	37.57 \\
	
	 + Random && 11.05 & 15.0 & 41.3& 38.05\\
	  + \textbf{Targeted} && \textbf{11.14}&\textbf{16.7}&	\textbf{41.24}&	\textbf{38.82} \\
	 
	\specialrule{1.2pt}{1pt}{1pt}

	\end{tabular}
	
	}
\caption{Targeting pedestrian detection models with position-specific neurons. We remove the top 100 neurons from the latent presentation of the detection models. Targeting the position-specific channels has more influence on the overall pedestrian detection performance compared to the random targeting. Note that lower is better for the reported metrics.}

    \label{tab:target_detection}
    \vspace{-0.2cm}
\end{table}

\section{Targeting Pedestrian Detection Networks} \label{sec:pedes}
We are interested in whether position-specific neurons are important for object-centric position-dependant tasks. Our hypothesis is that removing position-specific neurons may harm detection performance more than removing random neurons as position is an important factor in the successful detection of objects in a scene. To this end, we now target the overall position-specific channels of a pedestrian detection model trained on the CityPerson~\cite{cordts2016cityscapes} dataset. The CityPerson dataset is based on Cityscapes~\cite{cordts2016cityscapes} but only uses the bounding box annotations of the \textit{person} category and is used for the task of pedestrian detection. We choose the following three recent pedestrian detection models trained on CityPerson (available in~\cite{hasan2020pedestrian}): (i) Faster-RCNN~\cite{ren2015faster} (ii) Cascade-RCNN~\cite{cai2018cascade} with the HRNet~\cite{wang2020deep} backbone, and (iii) CSP-ResNet-50~\cite{liu2019high}. Similar to the experiment in Sec.~\ref{sec:target_segmentation}, we identify the top $N$ overall position-encoding channels (using Eq.~\ref{eq:rank_pos_abs}) and remove these dimensions before passing the latent representation to the detection head. 

Table~\ref{tab:target_detection} presents the pedestrian detection results when the top $100$ position-specific neurons are removed from a pedestrian detection model trained on CityPerson (we choose $N=100$ as the latent dimension of the networks used are relatively small (e.g., 256 for HRNet~\cite{wang2020deep})). Note that we follow the standard benchmark metric, mean average-precision (mAP), to report the detection results under four different settings. The results are consistent with the semantic segmentation results (Sec.~\ref{sec:target_segmentation}): removing the top 100 position-encoding channels degrades the performance more than choosing 100 random neurons. For example, for the Faster-RCNN network, targeting the position encoding neurons decreases the performance by 4.02\%, while targeting random neurons admits a 1.58\% drop.

\end{document}